\documentclass{article} 
\usepackage[preprint]{acl}

\usepackage{hyperref}
\usepackage{url}
\usepackage{booktabs}
\usepackage{glossaries}
\usepackage{graphicx}
\usepackage{tabularx}
\usepackage{float}

\glsdisablehyper

\newacronym[plural=LLMs,firstplural=Large Language Models (LLMs)]{llm}{LLM}{Large Language Model}
\newacronym{kd}{KD}{Knowledge Distillation}
\newacronym{sft}{SFT}{Supervised Fine-Tuning}
\newacronym{dpo}{DPO}{Direct Preference Optimization}
\newacronym{ppo}{PPO}{Proximal Policy Optimization}
\newacronym{rl}{RL}{Reinforcement Learning}
\newacronym{grpo}{GRPO}{Group Relative Policy Optimization}

\IfFileExists{_minted-main_v3/default.pygstyle}{}{}

\usepackage{amsmath}
\usepackage{amssymb}
\usepackage{mathtools}
\usepackage{amsthm}
\usepackage{placeins}
\usepackage[frozencache,cachedir=_minted-main_v3]{minted}
\usepackage{subcaption}

\usepackage{threeparttable}
\usepackage{dblfloatfix}

\usepackage{tikz}
\usetikzlibrary{arrows.meta}

\usepackage[capitalize,noabbrev]{cleveref}

\theoremstyle{plain}

\theoremstyle{definition}

\theoremstyle{remark}

\usepackage[textsize=tiny]{todonotes}

\title{torchtune: PyTorch native post-training library}

\author{
  \textbf{Mark Obozov\textsuperscript{1}},
  \textbf{Maxime Griot\textsuperscript{1,2}},
  \textbf{Joseph Cummings\textsuperscript{1}},
  \textbf{Evan Smothers\textsuperscript{1}},
  \\
  \textbf{Felipe Mello\textsuperscript{1}},
  \textbf{Rafi Ayub\textsuperscript{1}},
  \textbf{Philip John Bontrager\textsuperscript{1}},
  \textbf{Salman Mohammadi\textsuperscript{1}},
  \\
  \textbf{Ariel Kwiatkowski\textsuperscript{3}},
  \textbf{Nathan Azrak},
  \textbf{Mircea Mironenco},
\\
 \textsuperscript{1}Work done at PyTorch, Meta,
 \textsuperscript{2}Stanford,
 \textsuperscript{3}Meta-FAIR,
 \\
 \small{
    \textbf{Correspondence:} \href{mailto:obozovmark9@gmail.com}{obozovmark9@gmail.com}
  }
}

\begin{document}
\maketitle

\begin{abstract}
Modern \glspl{llm} typically require multistage training pipelines to achieve strong downstream performance, with post-training serving as the main interface for adapting open-weight models. We introduce \textbf{\texttt{torchtune}}, a PyTorch-native library designed to streamline the post-training lifecycle of \glspl{llm}, enabling efficient fine-tuning, experimentation, and deployment-oriented workflows. Unlike many existing fine-tuning frameworks, which often optimize for ease of use, specialized recipes, or hardware efficiency at the cost of transparency and extensibility, \texttt{torchtune} emphasizes modularity, hackability, and direct access to the underlying PyTorch components. In this paper, we present the design principles behind \texttt{torchtune}, describe how they are reflected in its model builders, training recipes, and distributed training stack, and evaluate the library across representative post-training settings. We compare against popular fine-tuning frameworks, including Axolotl and Unsloth, and show that \texttt{torchtune} provides strong performance and memory efficiency across many settings while remaining flexible enough for rapid research iteration. These results position \texttt{torchtune} as a practical foundation for reproducible \gls{llm} post-training research.
\end{abstract}

\title{torchtune: PyTorch native post-training library}

\begin{figure*}[t]
    \centering
    \resizebox{0.98\textwidth}{!}{%
    \begin{tikzpicture}[
        font=\small,
        node distance=3mm,
        command/.style={draw,rounded corners=2pt,fill=blue!8,minimum width=88mm,minimum height=10mm,align=center,inner sep=4pt,font=\footnotesize},
        cfg/.style={draw,rounded corners=2pt,fill=green!8,minimum width=31mm,minimum height=10mm,align=center,inner sep=3pt,font=\footnotesize},
        recipe/.style={draw,rounded corners=2pt,fill=gray!10,minimum width=48mm,minimum height=11mm,align=center,inner sep=4pt,font=\footnotesize},
        policy/.style={draw,rounded corners=2pt,fill=orange!10,minimum width=26mm,minimum height=10mm,align=center,inner sep=3pt,font=\footnotesize},
        resultbox/.style={draw,rounded corners=2pt,fill=red!10,text width=108mm,minimum height=12mm,align=center,inner sep=4pt,font=\footnotesize},
        paneltitle/.style={font=\scriptsize\bfseries,inner sep=1pt},
        >={Latex[length=2mm]}
    ]
        \node[command] (cmd) at (0,0) {\texttt{tune run} \emph{recipe} \texttt{--config} \emph{model/workflow} \texttt{key=value}\\\scriptsize choose a recipe skeleton, then override individual config fields};

        \draw[draw=gray!70,fill=gray!4,rounded corners=3pt] (-7.45,-1.05) rectangle (7.45,-4.05);
        \node[paneltitle] at (0,-1.3) {YAML config component graph};

        \node[cfg] (model) at (-5.5,-2.0) {model builder\\\scriptsize full / LoRA / QLoRA / ...};
        \node[cfg] (tok) at (-1.75,-2.0) {tokenizer\\\scriptsize Llama / Qwen / ... };
        \node[cfg] (data) at (1.75,-2.0) {dataset\\\scriptsize Alpaca / preference / ... };
        \node[cfg] (ckpt) at (5.5,-2.0) {checkpointer\\\scriptsize full / adapter};

        \node[cfg] (loss) at (-4.0,-3.35) {objective\\\scriptsize SFT CE / LCE / DPO / KD};
        \node[cfg] (optim) at (0,-3.35) {optimizer + scheduler\\\scriptsize AdamW / 8-bit / warmup};
        \node[cfg] (log) at (4.0,-3.35) {observability\\\scriptsize logger / profiler / metrics};

        \node[recipe] (setup) at (0,-4.95) {recipe setup\\\scriptsize \texttt{config.instantiate(...)} wires components and applies runtime policies};
        \node[recipe] (loop) at (0,-6.45) {recipe loop stays local and readable\\\scriptsize batch $\to$ forward $\to$ loss $\to$ backward $\to$ step $\to$ log};

        \draw[draw=gray!70,fill=orange!3,rounded corners=3pt] (-7.8,-9.1) rectangle (7.88,-7.45);
        \node[paneltitle] (runtime) at (0,-7.65) {runtime policies (applied at setup; wrap the loop)};

        \node[policy] (dist) at (-5.9,-8.4) {parallel\\\scriptsize FSDP2 / TP / CP / EP};
        \node[policy] (mem) at (-2.8,-8.4) {memory\\\scriptsize AC / offload};
        \node[policy] (compile) at (0.2,-8.4) {compile\\\scriptsize model / loss / optim};
        \node[policy] (stepfuse) at (3.2,-8.4) {step fusion\\\scriptsize Optim-in-bwd};
        \node[policy] (rl) at (6.2,-8.4) {RL orchestration\\\scriptsize sync / async GRPO};

        \node[resultbox] (result) at (0,-10.15) {\textbf{Swap one component or policy; keep the recipe shape.}\\\scriptsize The same source scales from 1 GPU to multi-gpu benchmarks.};

        \draw[->,thick] (0,-4.05) -- (setup.north);
        \draw[->,thick] (setup.south) -- (loop.north);
        \draw[->,thick,dashed] (0, -7.45) -- (loop.south);
    \end{tikzpicture}%
    }
    \caption{\textbf{Visual abstract.} \texttt{torchtune} recipes instantiate model, data, objective, optimizer, logging, and runtime components from YAML, then run them through a shared PyTorch training loop. Experiments are expressed by swapping component paths or runtime policies while preserving the recipe structure.}
    \label{fig:workflow}
\end{figure*}
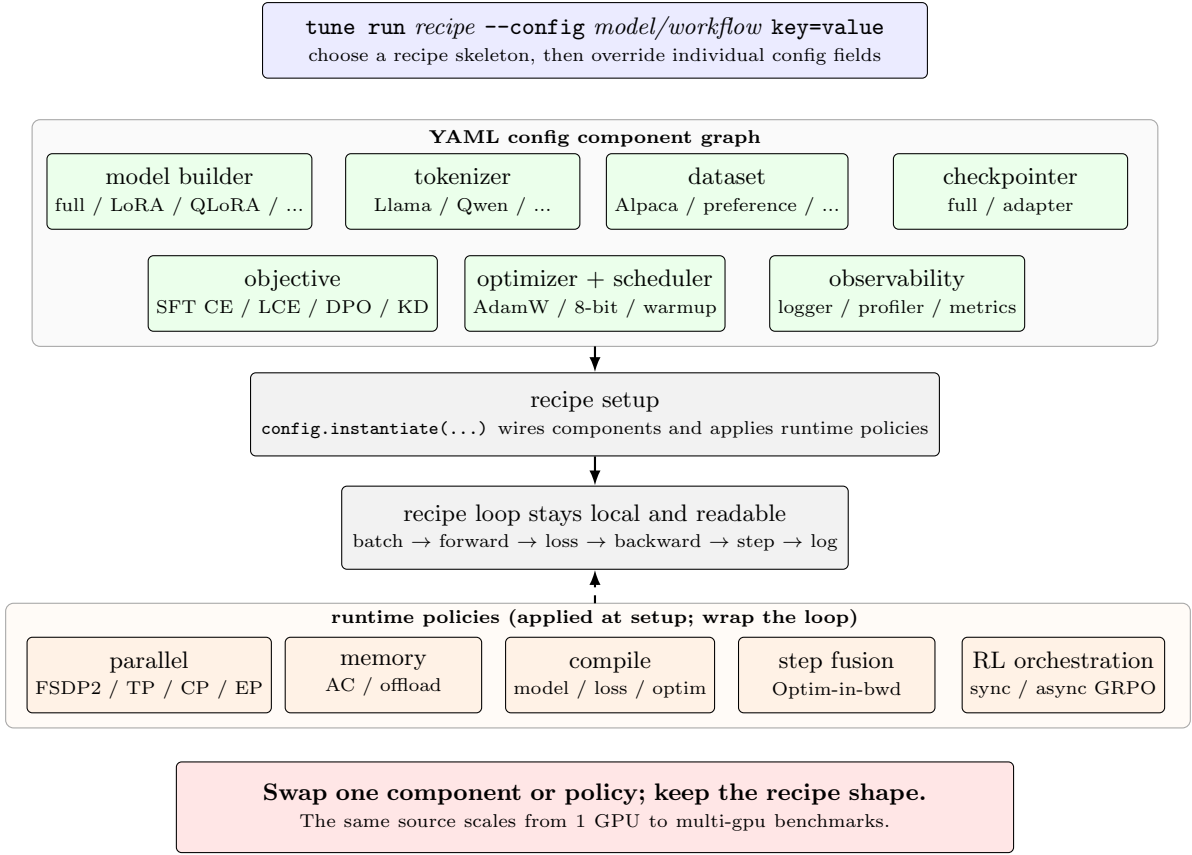

\section{Introduction}
\label{sec:intro}

Post-training, including supervised fine-tuning, preference optimization, distillation, and RL-based alignment, is now the dominant phase of effort for adapting open-weight \glspl{llm} to downstream tasks~\citep{lai-etal-2025-survey}. The frameworks that support this phase, however, have grown into complex stacks that make it hard to balance extensibility, performance, and rapid experimentation within a single tool.

Existing frameworks for \gls{llm} post-training include Axolotl~\citep{axolotl}, the Hugging Face \texttt{trl}/\texttt{transformers}/\texttt{peft} stack~\citep{vonwerra2022trl,peft}, and Unsloth~\citep{unsloth}. These systems occupy different points in the design space: Axolotl emphasizes configuration-driven fine-tuning, the Hugging Face ecosystem emphasizes broad integration and model coverage, and Unsloth emphasizes efficient LoRA~\citep{hu_lora_2022} and QLoRA~\citep{dettmers2023qlora} fine-tuning. Across these systems, several recurring tradeoffs motivate \texttt{torchtune}:

\begin{enumerate}
    \item \textbf{Large dependency stacks.} Frameworks built around \texttt{transformers} and adjacent libraries inherit broad transitive dependencies, which can complicate deployment, debugging, and reproducibility for focused post-training experiments.
    \item \textbf{Tightly coupled training decisions.} Model construction, trainer logic, distributed policy, and adapter insertion are often spread across factory layers or trainer abstractions, making fine-grained modifications harder than changing the underlying PyTorch modules directly.
    \item \textbf{Uneven access to modern PyTorch performance paths.} Generic implementations can leave memory and throughput on the table relative to FSDP2~\cite{fsdp-rfc}, DTensor~\citep{pytorch_contributors_torchdistributedtensor_2026}, loss parallelism, and \texttt{torch.compile}; kernel-specialized systems improve some paths but can make the training loop less transparent.
    \item \textbf{Fragmented multi-recipe support.} \gls{sft}, \gls{dpo}, \gls{ppo}/\gls{grpo}, knowledge distillation, and quantization-aware training often live in separate libraries, recipe families, or adapter systems, which makes controlled comparison difficult.
    \item \textbf{Distributed-training composability.} Multi-node training, tensor parallelism, and context parallelism are supported unevenly across frameworks, often requiring different backends or different abstractions at different scales.
\end{enumerate}

We introduce \texttt{torchtune}, a PyTorch~\citep{pytorch}-native post-training library designed around composable building blocks rather than monolithic trainers. \texttt{torchtune} provides hackable model builders, YAML-driven recipes inspired by Hydra~\citep{Yadan2019Hydra}, and a small set of orthogonal optimization switches whose effect on memory and throughput can be measured independently and combined freely. The same components scale from a single H100 to multi-node FSDP2 clusters without rewriting the training loop.

The contributions of this paper are:

\begin{enumerate}
    \item A \textbf{modular component model} (\S\ref{sec:design}) that makes model variants, LoRA, quantization, and architecture changes local to explicit PyTorch builders.
    \item An \textbf{in-backward optimizer fusion} for LLM training (\S\ref{sec:optim_bwd}) that reduces gradient-buffer lifetime and improves memory feasibility for large-model fine-tuning.
    \item A \textbf{composable parallelism stack} (\S\ref{sec:parallelism}) built on PyTorch DTensor: FSDP2, tensor and sequence parallel, expert parallel for MoE, loss parallel, and context parallel~\citep{pytorch_contributors_torchdistributedtensor_2026}.
    \item An \textbf{asynchronous GRPO} recipe (\S\ref{sec:async_grpo}) that decouples rollout generation from policy updates through a Ray-coordinated queue and replay buffer, supporting both on-policy synchronization and bounded off-policy lag.
    \item An \textbf{empirical study} (\S\ref{sec:perf}) comparing \texttt{torchtune} against Axolotl and Unsloth across single- and multi-GPU post-training settings.
\end{enumerate}

Our empirical sections quantify the optimization choices used throughout the evaluation. Across single-GPU and multi-GPU sweeps Llama~3.3~70B training, we isolate the effects of compilation, activation checkpointing~\citep{DBLP:journals/corr/ChenXZG16}, Linear Cross-Entropy, sequence packing, low-bit optimizer state, and in-backward optimizer fusion. The results show that these optimizations are complementary: compilation is a strong throughput lever on small and mid-sized models, and memory-oriented techniques determine feasibility at larger scales.

The code of the presented framework is available in the public \texttt{torchtune} repository.\footnote{\url{https://github.com/pytorch/torchtune}}

\section{Related Work}

\paragraph{General-purpose training frameworks.}
PyTorch Lightning~\citep{Falcon_PyTorch_Lightning_2019} and DeepSpeed~\citep{deepspeed} are among the most widely adopted systems for large-scale training. Lightning abstracts the training loop into a high-level Trainer API, enabling modularity and rapid prototyping across a variety of tasks. DeepSpeed focuses on scaling efficiency through the ZeRO family of optimizers~\citep{zero9355301}, supporting trillion-parameter models via optimizer and activation partitioning. Both frameworks are general in scope and not tailored to the specific challenges of LLM post-training.

\paragraph{Hugging Face ecosystem.}
The Hugging Face stack provides specialized libraries for adapting pretrained models. The \texttt{trl} library implements \gls{sft}, \gls{dpo}~\citep{rafailov2023direct}, \gls{ppo}, and reinforcement learning pipelines such as \gls{grpo}~\citep{vonwerra2022trl}, built atop the \texttt{transformers}~\citep{wolf-etal-2020-transformers} and \texttt{accelerate} libraries. Complementary to this, the \texttt{peft} library~\citep{peft} introduces parameter-efficient fine-tuning methods such as LoRA and QLoRA~\citep{dettmers2023qlora}. These frameworks emphasize ease of use and integration within the Hugging Face model hub, but rely heavily on the \texttt{transformers} training abstractions.

\paragraph{Community-driven frameworks.}
\texttt{Axolotl}~\citep{axolotl} has emerged as a community-oriented alternative for fine-tuning LLMs, exposing YAML-based configurations and extensive integration with Hugging Face libraries, PEFT adapters, and distributed runtimes such as DeepSpeed and DDP. Axolotl lowers the barrier for users by providing numerous preconfigured templates, but inherits the complexity of the underlying stack. \texttt{Unsloth}~\citep{unsloth} represents another line of work, targeting maximal efficiency for consumer-grade hardware through custom CUDA/Triton kernels~\citep{triton} that accelerate LoRA and QLoRA fine-tuning by up to $5\times$ while reducing memory usage. While effective, Unsloth prioritizes throughput and memory optimization over training loop transparency.

\paragraph{Position of \texttt{torchtune}.}
In contrast, \texttt{torchtune} is explicitly designed for the post-training lifecycle of LLMs. Rather than introducing new training abstractions, it provides transparent PyTorch-native recipes for \gls{sft}~\citep{ouyang2022training}, \gls{kd}~\citep{hinton2015distilling}, \gls{rl} variants (\gls{dpo}, \gls{ppo}~\citep{schulman2017proximal}, \gls{grpo}), and quantization-aware training. Scaling is achieved through native PyTorch FSDP, with memory-saving techniques integrated into reference configurations. Moreover, \texttt{torchtune} interoperates with Hugging Face Hub, LM Evaluation Harness~\citep{eval-harness}, and TorchAO~\citep{torchao}, while maintaining hackability through lightweight, modifiable scripts. This places it between high-level automated trainers and low-level performance-specialized kernels.

\section{Design}
\label{sec:design}

The \texttt{torchtune} library is designed around a modular architecture that promotes code reuse and experimentation. \cref{fig:workflow} summarizes how the pieces compose. The library provides low-level Transformer building blocks, model builders that assemble those blocks into complete architectures, and training recipes that orchestrate optimization, data loading, checkpointing, and distributed execution. The goal is not to hide PyTorch behind a new trainer abstraction, but to keep the research surface close to the code that is actually executed.

\subsection{Modular components and builders}
\label{sec:components_builders}

At the model level, \texttt{torchtune} keeps module assembly explicit: core Transformer blocks accept pre-constructed submodules, while builders choose the concrete attention, feed-forward, normalization, embedding, and projection implementations for each architecture. This makes architecture variants local to construction time: LoRA adapters, quantized projections, alternative attention kernels, or architecture-specific blocks can be swapped without rewriting the shared decoder logic or training recipe.

Builders also define the boundary for controlled experimentation. The same recipe can compare full fine-tuning, LoRA, QLoRA, quantization-aware training, or custom module replacements while keeping data, optimizer, logging, and distributed policy fixed. \cref{app:component_examples} provides abbreviated code examples for the decoder, attention module, builder loop, and LoRA substitution.

\subsection{Out-of-the-box support}

By removing the dependency on the \texttt{transformers} framework, \texttt{torchtune} ships pure-PyTorch reference implementations of modern open-source LLMs that are numerically equivalent with their original \texttt{transformers} counterparts but substantially simpler to read and modify. The list of supported model families is given in \cref{tab:models_support}; adding a new architecture amounts to writing a builder function in the style of \cref{sec:design}. \texttt{torchtune} also ships built-in dataset adapters for common post-training schemas such as chat, instruct, multimodal chat, preference (DPO), and text completion, listed in \cref{tab:datasets_support}, so user-curated datasets can be plugged in by selecting the matching adapter rather than writing custom collation code.

\paragraph{Other supported workflows.}
Beyond the \gls{sft}, \gls{dpo}, and \gls{grpo} recipes that this paper benchmarks, the same component model is also used to support \emph{multimodal} fine-tuning (early- and deep-fusion vision-language models such as Llama~3.2-Vision), \emph{knowledge distillation} with dedicated KD losses, and \emph{quantization-aware training} via TorchAO~\citep{torchao}.

\subsection{Training Recipes}

The recipe system is inspired by the Hydra configuration framework \citep{Yadan2019Hydra}: each recipe implements a training procedure (e.g.\ SFT, DPO, GRPO) and is parameterized by a YAML configuration. A representative default configuration for full single-device fine-tuning of Llama~3.1~8B is shown in \cref{app:config}.

Each block (model, dataset, optimizer, loss, logging) is independently swappable; an ablation that replaces \texttt{LinearCrossEntropyLoss} with the standard CE, or \texttt{PagedAdamW8bit} with \texttt{torch.optim.AdamW}, requires editing one line and changes nothing else in the recipe. Command-line overrides (e.g.\ \texttt{compile=True}) compose with the YAML for sweep-style experimentation.

\section{Implementation}
\label{sec:impl}

\subsection{In-backward optimizer fusion}
\label{sec:optim_bwd}

\paragraph{Motivation.}
Conventional optimizers for LLM training follow a two-phase procedure: gradients are first accumulated and stored for all parameters, and then consumed in a separate optimizer step. This requires a full gradient buffer $G_t = \{\nabla_{\theta_i} \mathcal{L}\}$ to remain in memory until the optimizer update is completed, adding significant memory overhead in addition to activations and optimizer states.

\paragraph{Design.}
\texttt{torchtune} introduces \emph{in-backward optimizer fusion}, in which the optimizer update is performed \emph{as part of the backward pass}. When the gradient for a parameter $\theta_i$ becomes available, it is immediately consumed by the optimizer (e.g., AdamW) to update both its state $(m_i, v_i)$ and the parameter value. The gradient can then be released without being retained for a global step. This design reduces the lifetime of gradient tensors from an entire backward pass to a single update site.

\paragraph{Implementation.}
Concretely, \texttt{torchtune}'s \texttt{OptimizerInBackward} wrapper instantiates one optimizer object per parameter and registers a \texttt{register\_post\_accumulate\_grad\_hook} on each parameter that calls \texttt{step()} and \texttt{zero\_grad()} as soon as the gradient is finalized. Per-parameter optimizer state is preserved separately and reassembled in the wrapper's \texttt{state\_dict} for checkpointing. Under one optimizer update per backward pass ($K{=}1$), the resulting parameter trajectory is bitwise identical to standard AdamW: the per-parameter updates commute because each $\theta_i$ has its own state $(m_i, v_i)$ with no cross-parameter dependency in AdamW's update rule. Building the optimizer step into the backward pass is itself not new at the PyTorch primitive level; our contribution is integrating it into an LLM post-training recipe and characterizing its memory and throughput effects across single-GPU and multi-GPU 70B regimes (\S\ref{sec:perf}).

\paragraph{Benefits.}
The fused strategy provides two main advantages:
\begin{itemize}
    \item \textbf{Memory efficiency:} peak gradient memory is reduced because gradients are consumed and discarded immediately. In our experiments this is the difference between OOM and a feasible run for Llama~3.3~70B on 8 H100s.
    \item \textbf{Computation overlap:} part of the optimizer workload is overlapped with backpropagation, which can yield modest throughput gains.
\end{itemize}

\paragraph{Limitations.}
\begin{itemize}
    \item \textbf{Gradient accumulation:} the method assumes one optimizer update per backward pass. When gradients must be accumulated over $K>1$ micro-batches, partial updates would be applied prematurely; benchmarks therefore compensate by raising the per-step batch size.
    \item \textbf{Distributed training:} the wrapper composes naturally with FSDP2 because each rank holds the relevant per-parameter shard, but it requires care to combine with optimizer-state-sharded ZeRO-style~\citep{deepspeed} schemes that themselves assume a global step.
\end{itemize}

\subsection{Linear Cross-Entropy loss}
\label{sec:lce}

\paragraph{Motivation and design.}
The final cross-entropy step in LLM training is a frequent peak-memory hotspot: materializing logits of shape $[B, S, V]$ can dominate memory for large vocabularies. Following the motivation of Cut Cross-Entropy~\citep{wijmans2025cut}, \texttt{torchtune}'s \texttt{LinearCrossEntropyLoss} fuses the final output projection with the cross-entropy computation, masks ignored tokens before projection, and processes hidden states in chunks so that the dense $[B, S, V]$ tensor is never materialized.

\paragraph{Composition and effect.}
Under tensor parallelism, the loss composes with PyTorch's loss-parallel context by sharding the output projection over the vocabulary dimension (\S\ref{sec:parallelism}). In our sweeps, LCE is not usually the dominant throughput lever, but it reliably reduces peak memory during loss computation and helps larger configurations compose with compilation and \texttt{Optim Bwd} (\S\ref{sec:perf}).

\section{Parallelism}
\label{sec:parallelism}

\texttt{torchtune} leverages PyTorch's newer DTensor API for distributed training, allowing seamless integration of multiple parallelism setups.

\textbf{Data parallelism} uses the \textbf{FSDPv2} APIs~\citep{zhao2023pytorchfsdpexperiencesscaling}. The data parallelism involves a 2D mesh, supporting data parallel replicates in addition to sharding, for maximizing use of high-bandwidth intra-node communication while reducing inter-node communication.

For \textbf{tensor parallelism}, \texttt{torchtune} specifies tensor and sequence parallel plans for key model architectures, as well as a custom \textbf{expert parallel} plan for mixture of expert models. This allows efficient parallelisation of models too large to train only with data parallelism.

To reduce the peak memory usage most commonly found in the logit and softmax tensors during cross-entropy loss calculation, \texttt{torchtune} utilises PyTorch's loss parallel context manager, sharding the output features over the vocab dimension across the tensor parallelism mesh. This helps to avoid materializing full logits and helps to shard both memory and computation over the vocabulary dim. In addition to the custom linear cross entropy implementation, this allows training of models with very large vocabularies on inputs with long contexts.

Beyond the classical parallelism types, \texttt{torchtune} includes context parallelism via Ring Attention~\citep{Liu2023RingAW}. Context parallelism in \texttt{torchtune} is analyzed in Appendix A.

\subsection{Performance Benchmarks}
\label{sec:perf}

\begin{table}[t]
\centering
\small
\setlength{\tabcolsep}{6pt}
\renewcommand{\arraystretch}{1.05}
\resizebox{\columnwidth}{!}{%
\begin{threeparttable}
\begin{tabular}{@{}l|cc|cc@{}}
\toprule
 & \multicolumn{2}{c|}{Qwen3 32B} & \multicolumn{2}{c}{Llama 3.3 70B} \\
\midrule
 & Memory & Tok/s & Memory & Tok/s \\
\midrule
\texttt{torchtune}    & 67.78 & 465.79 & OOM   & OOM \\
+ AC         & 42.93 & 405.95 & 74.75 & 122.55 \\
+ LCE        & 38.43 & 398.56 & OOM   & OOM \\
+ Compile\tnote{a} & 44.17 & 433.12 & 75.22 & 128.57 \\
+ Optim Bwd  & 60.41 & \textbf{581.63} & 74.89 & \textbf{352.11} \\
\texttt{Axolotl}      & 40.9  & 218    & OOM   & OOM \\
\bottomrule
\end{tabular}

\begin{tablenotes}[flushleft]
\footnotesize
\item[a] For Llama 3.3 we do not compile the optimizer step.
\end{tablenotes}
\end{threeparttable}
}
\caption[Single-node 32B/70B scaling]{Single-node memory (GB/GPU) and throughput (tokens/s/GPU) for Qwen3~32B and Llama~3.3~70B on $1{\times}8$~H100s, FSDP2 with tensor, data, and loss parallelism enabled by default; Alpaca, sequence length 2048, micro-batch size 2 (16 with \texttt{Optim Bwd}).}
\label{tab:mem_throughput_comp}
\end{table}

In terms of performance, we compared \texttt{torchtune} with Axolotl and Unsloth. We perform multiple runs enabling sequentially the optimizations introduced by \texttt{torchtune}. To evaluate the diverse training settings we perform two types of training:

\begin{enumerate}
    \item \textbf{Single Device}: Small models are trained on a single H100 GPU to test the ideal training conditions without risking bottlenecks introduced by interconnection. For this run we evaluate both full weight training and Low Rank Adaptation (LoRA).
    \item \textbf{Multi Device}: Larger models up to 70b parameters are trained using FSDP on 8 H100 GPUs on a single node. This allows us to evaluate the impact of interconnection and algorithmic efficiency of tensor parallel implementations.
\end{enumerate}

For all training jobs except the preference optimization jobs we use the Alpaca dataset, a sequence length of 2048, micro batch size of 2, gradient accumulation of 8, and use the default AdamW torch optimizer. In case of the DPO post-training we use helpful subset of the Anthropic's helpful/harmless RLHF data. The only exceptions are runs which require optimizer in backward. In this case we set gradient accumulation to 1, compensating it with the batch size of 16. For LoRA runs we use $\alpha = 16$ and $r = 16$.

Single device runs are executed on a server with 8 NVIDIA H100 GPUs, for fairness we always use device 0. To demonstrate the impact of each optimization introduced in \texttt{torchtune}, we enable sequentially each one of them and observe the compounded effect as shown in \cref{tab:single_device_full}. For LoRA the \texttt{torchtune} run already has all the optimizations enabled, we also use the AdamW 8bits optimizer from bitsandbytes with Axolotl.

\begin{table*}[t]
\centering
\small
\setlength{\tabcolsep}{4pt}
\renewcommand{\arraystretch}{1.05}
\caption[Training ablations on Qwen3]{Single-GPU (1$\times$H100) full-weight and LoRA ablations on Qwen3~\citep{yang2025qwen3} 0.6B--8B; Alpaca, sequence length 2048, micro-batch size 2 (16 with \texttt{Optim Bwd}); LoRA $r{=}16$, $\alpha{=}16$. Memory in GB; throughput in tokens/s/GPU.}
\begin{tabular}{@{}l|cc|cc|cc|cc@{}}
\toprule
 & \multicolumn{2}{c|}{Qwen3 0.6B} & \multicolumn{2}{c|}{Qwen3 1.7B} &
   \multicolumn{2}{c|}{Qwen3 4B}   & \multicolumn{2}{c}{Qwen3 8B} \\
\midrule
\textbf{Full weight} & Memory & Tok/s & Memory & Tok/s & Memory & Tok/s & Memory & Tok/s \\
\midrule
\texttt{torchtune}     & 8.60 & 5213  & 18.48 & 8071  & 31.52 & 2801 & OOM   & OOM \\
+ LCE         & 8.41 & 5454  & 19.49 & 7806  & 31    & 3099 & OOM   & OOM \\
+ Compile     & 6.96 & 7872  & 15.2  & \textbf{9621} & 31 & \textbf{3755} & OOM & OOM \\
+ AC          & \textbf{5.47} & 4836 & 13.95 & 7568  & 30    & 2983 & 64.04 & 3037 \\
+ Optim Bwd   & 6.1  & \textbf{20384} & \textbf{11.65} & 6743 & \textbf{24.37} & 2523 & \textbf{51.79} & \textbf{3773} \\
+ AdamW8Bit   & 3.67 & 15369 & 4.92  & 3873  & 9.50  & 1366 & 31.07 & 3066 \\
\midrule
\texttt{Axolotl}       & 8.63 & 3300  & 18.4  & 3295  & 41.1  & 2691 & 76.6  & 1903 \\
\midrule
\textbf{LoRA} & Memory & Tok/s & Memory & Tok/s & Memory & Tok/s & Memory & Tok/s \\
\midrule
\texttt{torchtune}     & 2.6 & \textbf{3292} & 4.6 & \textbf{3610} & 9.3 & \textbf{2616} & 17.2 & \textbf{2745} \\
\texttt{Axolotl}       & 6.2 & 1973 & 9.4 & 2233 & 17.7 & 1605 & 27.9 & 1609 \\
\texttt{Unsloth}       & \textbf{2.3} & 2502 & \textbf{4.4} & 3284 & \textbf{8.9} & 1826 & \textbf{16.8} & 1836 \\
\bottomrule
\end{tabular}
\label{tab:single_device_full}
\end{table*}

\begin{table*}[t]
\centering
\small
\setlength{\tabcolsep}{4pt}
\renewcommand{\arraystretch}{1.05}
\begin{tabular}{@{}l|cc|cc|cc|cc@{}}
\toprule
 & \multicolumn{2}{c|}{Qwen3 0.6B} & \multicolumn{2}{c|}{Qwen3 1.7B} &
   \multicolumn{2}{c|}{Qwen3 4B}   & \multicolumn{2}{c}{Qwen3 8B} \\
\midrule
\textbf{Full weight} & Memory & Tok/s & Memory & Tok/s & Memory & Tok/s & Memory & Tok/s \\
\midrule
\texttt{torchtune} & 5.47 & 4,836 & 13.95 & 7,568 & 30 & 2,983 & 64.04 & 3,037 \\
+ Packed 2048  & 11.79   & 57,031   & 15.88 & 29,466 &  33.92  & 13,538   & 76.41  & \textbf{8094}  \\
+ Packed 4096  & 18.37   & \textbf{85,924}   & 17.81 & \textbf{34,343} & 36.15   & 15,272   & OOM  & OOM  \\
\bottomrule
\end{tabular}
\caption[Packing on Qwen3]{Separate analysis of sequence packing on Qwen3 (single GPU). Memory in GB; throughput in tokens/s/GPU.}
\label{tab:single_device_packed}
\end{table*}

\begin{table}[t]
\centering
\small
\setlength{\tabcolsep}{6pt}
\renewcommand{\arraystretch}{1.05}
\resizebox{\columnwidth}{!}{%
\begin{tabular}{@{}l|cc|cc@{}}
\toprule
 & \multicolumn{2}{c|}{Qwen3 8B} & \multicolumn{2}{c}{Llama 3.1 8B} \\
\midrule
\textbf{Full weight} & Memory & Tok/s & Memory & Tok/s \\
\midrule
DPO (torchtune)   & 91.02 & \textbf{782.2} & 81.82 & \textbf{745.0} \\
DPO (axolotl)     & OOM & - & OOM & - \\
DPO (axolotl + 8-bit)     & 69.66 & 185.6 & 67.64 & 249.2 \\
\bottomrule
\end{tabular}
}
\caption[RLHF procedures]{Comparison of training speed and memory usage between \texttt{Axolotl} and \texttt{torchtune} on DPO. Experiments were conducted on single GH200 with 96GB HBM3 vRAM. \texttt{torchtune} was able to fit the training in memory with the standard AdamW optimizer. Still, due to out-of-memory errors with \texttt{Axolotl}, we also report runs with the 8-bit optimizer from bitsandbytes~\citep{dettmers2022optimizers}.}
\label{tab:single_device_rlhf}
\end{table}

\section{Analysis of applied optimizations}

\texttt{torchtune} exposes a set of composable training-time optimizations that primarily target (i) peak activation memory, (ii) optimizer-state memory, and (iii) kernel/graph efficiency. We evaluate these flags in a controlled single-GPU setting across Qwen3 model sizes (\cref{tab:single_device_full}) and for larger-scale configurations under parallelism (\cref{tab:mem_throughput_comp}). Overall, the results illustrate a consistent trade-off surface: techniques that reduce memory pressure often introduce recomputation or lower arithmetic intensity, while compiler- and kernel-level optimizations primarily improve throughput and can sometimes reduce memory indirectly by fusing operations and eliminating intermediates.

Across Qwen3-0.6B/1.7B/4B, enabling \texttt{compile}~\citep{ansel2024pytorch2} yields the most reliable throughput improvements, and in several cases also reduces peak memory relative to the baseline (\cref{tab:single_device_full}). For example, Qwen3-0.6B improves from 5.2k$\rightarrow$7.9k tok/s while reducing peak memory from 8.6$\rightarrow$7.0\,GB, and Qwen3-1.7B reaches its highest throughput under \texttt{compile} (9.6k tok/s). In contrast, \texttt{LCE} (chunked loss computation) provides comparatively modest and model-dependent gains: it can slightly improve throughput or stability for some settings, but it is not a dominant driver compared to compilation or activation rematerialization, while it can be useful for smoothing peak memory during the loss calculation step. Activation checkpointing (\texttt{AC}) is a dependable memory lever: it reduces peak memory across the board (notably enabling Qwen3-8B to run where the baseline configuration OOMs), at the expected cost of lower throughput.

For larger models where optimizer and gradient-state memory become prominent, two additional strategies materially expand feasibility (\cref{tab:mem_throughput_comp}). First, performing the optimizer step during the backward pass (\texttt{Optim Bwd}) can significantly improve throughput in memory-constrained regimes by reducing the lifetime of gradient buffers (while also imposing constraints such as limited compatibility with gradient accumulation and lack of improvement in cases when the gradients might be small: FSDP runs or models with few parameters). Second, low-bit optimizer state (\texttt{AdamW8Bit}) yields the largest absolute memory reductions in the Qwen3 sweep (e.g., Qwen3-1.7B: 11.7$\rightarrow$4.9\,GB) but may reduce throughput depending on the configuration. Notably, for Llama~3.3~70B the baseline configuration OOMs, while combinations that include \texttt{AC} enable training and \texttt{Optim Bwd} further increases throughput (\cref{tab:mem_throughput_comp}). 

Finally, we report sequence packing as a separate axis of analysis (\cref{tab:single_device_packed}), since it changes effective token utilization and interacts with the memory/throughput balance.

\section{Asynchronous GRPO}
\label{sec:async_grpo}

\gls{grpo}~\citep{shao2024deepseekmath} optimizes a policy using scalar rewards computed over groups of sampled generations. \texttt{torchtune} provides two distributed full fine-tuning recipes: \texttt{grpo\_full\_finetune\_distributed} (synchronous, closely following the original algorithm) and \texttt{async\_grpo\_full\_finetune\_distributed}, which overlaps generation and optimization to increase end-to-end accelerator utilization.

\paragraph{Motivation.}
In synchronous \gls{grpo}, decoding and optimization are serialized on the same set of devices, which can lead to under-utilization whenever one phase is significantly more expensive than the other (typically decoding for long rollouts). Asynchronous \gls{grpo} mitigates this by decoupling rollout generation from policy updates and executing both phases concurrently.

\begin{figure}[t]
    \centering
    \begin{subfigure}{\linewidth}
        \centering
        \resizebox{\linewidth}{!}{%
        \begin{tikzpicture}[
            x=0.75cm,
            y=0.85cm,
            box/.style={draw,rounded corners=1pt,minimum height=0.55cm,inner sep=2pt,font=\scriptsize,align=center},
            g/.style={box,fill=blue!7},
            t/.style={box,fill=orange!10},
            s/.style={box,fill=gray!10},
            >={Latex[length=2mm]}
        ]
            \draw[->,thin] (0,0) -- (12,0) node[right,font=\scriptsize]{time};

            \node[g,minimum width=2cm] (g1) at (1.3,0.9) {generate};
            \node[s,right=0cm of g1] (sh1) {shard};
            \node[t,right=0cm of sh1,minimum width=1.5cm] (t1) {train};
            \node[s,right=0cm of t1] (sh2) {shard};
            \node[g,right=0cm of sh2,minimum width=2cm] (g2) {generate};
            \node[s,right=0cm of g2] (sh3) {shard};
            \node[t,right=0cm of sh3,minimum width=1.5cm] (t2) {train};

        \end{tikzpicture}%
        }
        \caption{Synchronous alternation of generation and training.}
        \label{fig:grpo_sched_sync}
    \end{subfigure}

    \vspace{4pt}

    \begin{subfigure}{\linewidth}
        \centering
        \resizebox{\linewidth}{!}{%
        \begin{tikzpicture}[
            x=0.78cm,
            y=0.9cm,
            box/.style={draw,rounded corners=1pt,minimum height=0.55cm,inner sep=2pt,font=\scriptsize,align=center},
            g/.style={box,fill=blue!7},
            t/.style={box,fill=orange!10},
            sync/.style={box,fill=gray!10},
            lane/.style={font=\scriptsize,anchor=east},
            >={Latex[length=2mm]}
        ]
            \draw[->,thin] (0,-0.8) -- (11.0,-0.8) node[right,font=\scriptsize]{time};

            \node[lane] at (-0.2,1.2) {generator};
            \node[lane] at (-0.2,0.2) {trainer};

            \node[g,anchor=west,minimum width=2.5cm] (g1) at (0.0,1.2) {generate};
            \node[g,anchor=west,minimum width=2.5cm] (g2) at (5.15,1.2) {generate};
            \node[g,anchor=west,minimum width=1cm] (g3) at (10.15,1.2) {generate};

            \node[t,anchor=west,minimum width=2.5cm] (t1) at (0.5,0.2) {train};
            \node[sync,anchor=west,minimum width=1.1cm,right=0cm of t1] (s1) {sync};

            \node[t,anchor=west,minimum width=2.5cm] (t2) at (5.5,0.2) {train};
            \node[sync,anchor=west,minimum width=1.1cm,right=0cm of t2] (s2) {sync};

            \draw[-,thick] (s1.south east) -- (s1.north east |- g2.north);
            \draw[-,thick] (s2.south east) -- (s2.north east |- g3.north);

        \end{tikzpicture}%
        }
        \caption{On-policy asynchronous overlap between decoding and optimization.}
        \label{fig:grpo_sched_async_onpolicy}
    \end{subfigure}

    \vspace{2pt}

    \begin{subfigure}{\linewidth}
        \centering
        \resizebox{\linewidth}{!}{%
        \begin{tikzpicture}[
            x=0.78cm,
            y=0.9cm,
            box/.style={draw,rounded corners=1pt,minimum height=0.55cm,inner sep=2pt,font=\scriptsize,align=center},
            g/.style={box,fill=blue!7},
            t/.style={box,fill=orange!10},
            sync/.style={box,fill=gray!10},
            lane/.style={font=\scriptsize,anchor=east},
            >={Latex[length=2mm]}
        ]
            \draw[->,thin] (0,-0.8) -- (11.0,-0.8) node[right,font=\scriptsize]{time};

            \node[lane] at (-0.2,1.2) {generator};
            \node[lane] at (-0.2,0.2) {trainer};

            \node[g,anchor=west,minimum width=3cm] (g1) at (0.0,1.2) {generate};
            \node[g,anchor=west,minimum width=3cm,right=0cm of g1] (g2) {generate};
            \node[g,anchor=west,minimum width=3cm,right=0cm of g2] (g3) {generate};

            \node[t,anchor=west,minimum width=1.5cm] (t1) at (0.5,0.2) {train};
            \node[sync,anchor=west,minimum width=1.1cm,right=0cm of t1] (s1) {sync};

            \node[t,anchor=west,minimum width=1.5cm] (t2) at (4.4,0.2) {train};
            \node[sync,anchor=west,minimum width=1.08cm,right=0cm of t2] (s2)  {sync};

            \node[t,anchor=west,minimum width=1.5cm] (t3) at (8.3,0.2) {train};
            \node[sync,anchor=west,minimum width=1.08cm,right=0cm of t3] (s3)  {sync};

            \draw[-,thick] (s1.south east) -- (s1.north east |- g2.north);
            \draw[-,thick] (s2.south east) -- (s2.north east |- g3.north);

        \end{tikzpicture}%
        }
        \caption{Controlled off-policy rollouts, trading policy lag for smoother training throughput.}
        \label{fig:grpo_sched_async_offpolicy}
    \end{subfigure}

    \caption{Timeline schematics for \gls{grpo} execution. ``sync'' denotes a parameter refresh from trainer to generator; ``shard'' denotes sharded-state transitions associated with distributed training.}
    \label{fig:grpo_schedules}
\end{figure}
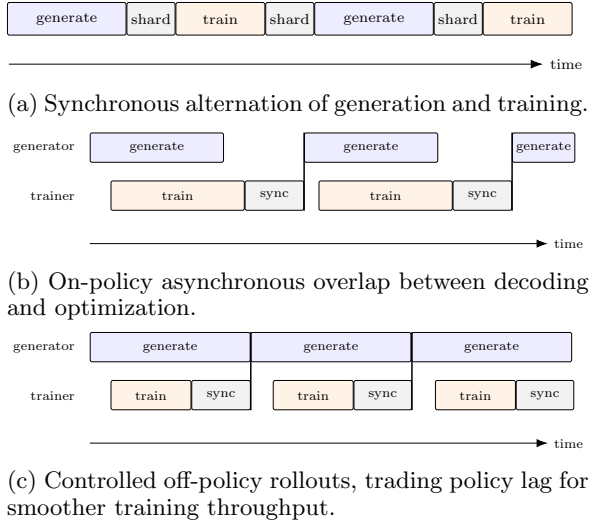

\paragraph{Scheduling regimes.}
Figure~\ref{fig:grpo_schedules} summarizes three scheduling regimes supported by our implementation: (i) a synchronous baseline, (ii) an on-policy asynchronous pipeline with periodic synchronization, and (iii) a controlled off-policy variant that permits bounded policy lag in exchange for higher utilization.

\paragraph{System architecture and design choices.}
The asynchronous recipe is implemented as a set of Ray-coordinated workers~\citep{moritz2018ray}, with a deliberate separation between (i) the GRPO algorithmic computation and (ii) orchestration and systems concerns.

\textbf{Orchestration as a recipe.}
We implement the distributed system as an orchestration recipe (\texttt{OrchestrationRecipeInterface}) whose responsibility is to instantiate and wire reusable worker roles (generation, post-processing, training, logging). This keeps GRPO math encapsulated inside existing \texttt{torchtune} workers and makes it straightforward to change execution models (synchronous vs.
async) without rewriting the learning algorithm.

\textbf{Decoupling via a queue + replay buffer.}
Rollouts are moved through two explicit data structures: a bounded Ray \texttt{Queue} used for low-latency handoff and backpressure control between rollout workers and post-processing workers, and a \texttt{torchrl} \texttt{RayReplayBuffer}~\citep{bou2024torchrl} used to store processed trajectories for training. The queue controls burstiness; the replay buffer provides a uniform sampling API that naturally supports controlled policy lag.

\textbf{Worker roles.}
\begin{itemize}
    \item \textbf{Inference (rollout workers):} vLLM-backed collectors (\texttt{SyncLLMCollector}) sample grouped responses and token-level log-probabilities and push raw rollouts into the shared queue.
    \item \textbf{Post-processing (reference workers):} \texttt{PostProcessingWorker} actors pop rollouts, compute GRPO-specific training fields (e.g., rewards, masks, and per-trajectory metadata), and write processed trajectories to the replay buffer.
    \item \textbf{Training (actor workers):} distributed \texttt{TrainingWorker} actors consume mini-batches from the replay buffer and apply policy updates.
\end{itemize}

\textbf{Weight synchronization.}
Training runs under PyTorch distributed (FSDP), while generation runs in vLLM~\citep{kwon2023efficient} with optional tensor parallelism. We connect these worlds through a dedicated vLLM parameter server that the training workers can publish to, and which the rollout workers poll via lightweight weight-update receivers. This design refreshes decoding weights online without restarting collectors.

\textbf{Observability and concurrency knobs.}
To make asynchronous behavior debuggable, we inject a single \texttt{MetricLoggerWorker} across all roles for consistent reporting of rollout latency, queue/backpressure behavior, replay-buffer throughput, and reward statistics; and we expose key concurrency/resource choices (e.g., queue max size, replay-buffer capacity, and Ray \texttt{max\_concurrency}) in configuration.

\paragraph{End-to-end flow.}
At runtime, (i) the generator produces grouped rollouts (prompt, sampled responses, and associated log-probabilities), (ii) the trainer augments trajectories with reward and optimization-specific fields (reference log-probabilities, advantages, and masks), and (iii) the trainer updates the policy and triggers parameter synchronization at a configurable cadence.

\paragraph{Scope of evaluation.}
We present asynchronous GRPO as a \emph{system-design contribution}: the recipe separates GRPO computation from orchestration concerns (queueing, replay, weight synchronization) while exposing both on-policy and bounded-lag execution in configuration. We leave head-to-head reward and throughput comparisons to future work, since they are sensitive to reward design, sampling stochasticity, and policy-lag tuning. The recipe (\texttt{async\_grpo\_full\_finetune\_distributed}) is included in the released code.

\section{Conclusion}
\label{sec:conclusion}

We presented \texttt{torchtune}, a PyTorch-native library for the post-training lifecycle of \glspl{llm} that prioritizes modularity, transparency, and hardware efficiency. By organizing post-training around composable model builders, data pipelines, recipes, and optimization choices, \texttt{torchtune} enables rapid experimentation without forcing users into rigid abstractions. Across the supported model and dataset interfaces (\cref{tab:models_support,tab:datasets_support}), the library provides practical defaults while remaining extensible for custom architectures and data formats.

Our evaluation shows that \texttt{torchtune} can achieve competitive or superior efficiency compared with widely used post-training frameworks while maintaining an implementation that is easy to inspect and adapt. The experiments also show that the optimization choices are complementary: compilation improves throughput in many small and mid-sized settings, memory-oriented techniques often determine whether larger configurations are feasible, and linear cross-entropy reduces peak memory during loss computation. Together, these results position \texttt{torchtune} as a practical foundation for reproducible post-training studies that balance clarity, extensibility, and performance.

%
%
%

\FloatBarrier

\clearpage

\bibliography{references}

\begin{thebibliography}{33}
\providecommand{\natexlab}[1]{#1}

\bibitem[{Ansel et~al.(2024)Ansel, Yang, He, Gimelshein, Jain, Voznesensky, Bao, Bell, Berard, Burovski, Chauhan, Chourdia, Constable, Desmaison, DeVito, Ellison, Feng, Gong, Gschwind, Hirsh, Huang, Kalambarkar, Kirsch, Lazos, Lezcano, Liang, Liang, Lu, Luk, Maher, Pan, Puhrsch, Reso, Saroufim, Siraichi, Suk, Zhang, Suo, Tillet, Zhao, Wang, Zhou, Zou, Wang, Mathews, Wen, Chanan, Wu, and Chintala}]{ansel2024pytorch2}
Jason Ansel, Edward Yang, Horace He, Natalia Gimelshein, Animesh Jain, Michael Voznesensky, Bin Bao, Peter Bell, David Berard, Evgeni Burovski, Geeta Chauhan, Anjali Chourdia, Will Constable, Alban Desmaison, Zachary DeVito, Elias Ellison, Will Feng, Jiong Gong, Michael Gschwind, and 30 others. 2024.
\newblock \href {https://doi.org/10.1145/3620665.3640366} {Pytorch 2: Faster machine learning through dynamic python bytecode transformation and graph compilation}.
\newblock In \emph{Proceedings of the 29th ACM International Conference on Architectural Support for Programming Languages and Operating Systems, Volume 2}, ASPLOS '24, page 929–947, New York, NY, USA. Association for Computing Machinery.

\bibitem[{{Axolotl maintainers and contributors}(2023)}]{axolotl}
{Axolotl maintainers and contributors}. 2023.
\newblock \href {https://github.com/axolotl-ai-cloud/axolotl} {Axolotl: Open source llm post-training}.

\bibitem[{Bou et~al.(2024)Bou, Bettini, Dittert, Kumar, Sodhani, Yang, De~Fabritiis, and Moens}]{bou2024torchrl}
Albert Bou, Matteo Bettini, Sebastian Dittert, Vikash Kumar, Shagun Sodhani, Xiaomeng Yang, Gianni De~Fabritiis, and Vincent Moens. 2024.
\newblock \href {https://proceedings.iclr.cc/paper_files/paper/2024/file/07bc8125400bf4b140c332010756bd9b-Paper-Conference.pdf} {Torchrl: A data-driven decision-making library for pytorch}.
\newblock In \emph{International Conference on Learning Representations}, volume 2024, pages 1778--1811.

\bibitem[{Chen et~al.(2016)Chen, Xu, Zhang, and Guestrin}]{DBLP:journals/corr/ChenXZG16}
Tianqi Chen, Bing Xu, Chiyuan Zhang, and Carlos Guestrin. 2016.
\newblock \href {https://arxiv.org/abs/1604.06174} {Training deep nets with sublinear memory cost}.
\newblock \emph{CoRR}, abs/1604.06174.

\bibitem[{Daniel~Han and team(2023)}]{unsloth}
Michael~Han Daniel~Han and Unsloth team. 2023.
\newblock \href {http://github.com/unslothai/unsloth} {Unsloth}.

\bibitem[{Dettmers et~al.(2022)Dettmers, Lewis, Shleifer, and Zettlemoyer}]{dettmers2022optimizers}
Tim Dettmers, Mike Lewis, Sam Shleifer, and Luke Zettlemoyer. 2022.
\newblock 8-bit optimizers via block-wise quantization.
\newblock \emph{9th International Conference on Learning Representations, ICLR}.

\bibitem[{Dettmers et~al.(2023)Dettmers, Pagnoni, Holtzman, and Zettlemoyer}]{dettmers2023qlora}
Tim Dettmers, Artidoro Pagnoni, Ari Holtzman, and Luke Zettlemoyer. 2023.
\newblock \href {https://proceedings.neurips.cc/paper_files/paper/2023/file/1feb87871436031bdc0f2beaa62a049b-Paper-Conference.pdf} {Qlora: Efficient finetuning of quantized llms}.
\newblock In \emph{Advances in Neural Information Processing Systems}, volume~36, pages 10088--10115. Curran Associates, Inc.

\bibitem[{Falcon and {The PyTorch Lightning team}(2019)}]{Falcon_PyTorch_Lightning_2019}
William Falcon and {The PyTorch Lightning team}. 2019.
\newblock \href {https://doi.org/10.5281/zenodo.3828935} {{PyTorch Lightning}}.

\bibitem[{Gao et~al.(2024)Gao, Tow, Abbasi, Biderman, Black, DiPofi, Foster, Golding, Hsu, Le~Noac'h, Li, McDonell, Muennighoff, Ociepa, Phang, Reynolds, Schoelkopf, Skowron, Sutawika, Tang, Thite, Wang, Wang, and Zou}]{eval-harness}
Leo Gao, Jonathan Tow, Baber Abbasi, Stella Biderman, Sid Black, Anthony DiPofi, Charles Foster, Laurence Golding, Jeffrey Hsu, Alain Le~Noac'h, Haonan Li, Kyle McDonell, Niklas Muennighoff, Chris Ociepa, Jason Phang, Laria Reynolds, Hailey Schoelkopf, Aviya Skowron, Lintang Sutawika, and 5 others. 2024.
\newblock \href {https://doi.org/10.5281/zenodo.12608602} {The language model evaluation harness}.

\bibitem[{Hinton et~al.(2015)Hinton, Vinyals, and Dean}]{hinton2015distilling}
Geoffrey Hinton, Oriol Vinyals, and Jeffrey Dean. 2015.
\newblock \href {http://arxiv.org/abs/1503.02531} {Distilling the knowledge in a neural network}.
\newblock In \emph{NIPS Deep Learning and Representation Learning Workshop}.

\bibitem[{Hu et~al.(2022)Hu, shen, Wallis, Allen-Zhu, Li, Wang, Wang, and Chen}]{hu_lora_2022}
Edward~J. Hu, yelong shen, Phillip Wallis, Zeyuan Allen-Zhu, Yuanzhi Li, Shean Wang, Lu~Wang, and Weizhu Chen. 2022.
\newblock \href {https://openreview.net/forum?id=nZeVKeeFYf9} {{LoRA}: {Low}-{Rank} {Adaptation} of {Large} {Language} {Models}}.
\newblock In \emph{International {Conference} on {Learning} {Representations}}.

\bibitem[{Kwon et~al.(2023)Kwon, Li, Zhuang, Sheng, Zheng, Yu, Gonzalez, Zhang, and Stoica}]{kwon2023efficient}
Woosuk Kwon, Zhuohan Li, Siyuan Zhuang, Ying Sheng, Lianmin Zheng, Cody~Hao Yu, Joseph~E. Gonzalez, Hao Zhang, and Ion Stoica. 2023.
\newblock Efficient memory management for large language model serving with pagedattention.
\newblock In \emph{Proceedings of the ACM SIGOPS 29th Symposium on Operating Systems Principles}.

\bibitem[{Lai et~al.(2025)Lai, Liu, Gao, Cheng, Qi, Xu, Yao, Zhang, Du, Hou, Lv, Huang, Dong, and Tang}]{lai-etal-2025-survey}
Hanyu Lai, Xiao Liu, Junjie Gao, Jiale Cheng, Zehan Qi, Yifan Xu, Shuntian Yao, Dan Zhang, Jinhua Du, Zhenyu Hou, Xin Lv, Minlie Huang, Yuxiao Dong, and Jie Tang. 2025.
\newblock \href {https://doi.org/10.18653/v1/2025.acl-long.140} {A survey of post-training scaling in large language models}.
\newblock In \emph{Proceedings of the 63rd Annual Meeting of the Association for Computational Linguistics (Volume 1: Long Papers)}, pages 2771--2791, Vienna, Austria. Association for Computational Linguistics.

\bibitem[{Liu et~al.(2023)Liu, Zaharia, and Abbeel}]{Liu2023RingAW}
Hao Liu, Matei Zaharia, and Pieter Abbeel. 2023.
\newblock \href {https://api.semanticscholar.org/CorpusID:263608461} {Ring attention with blockwise transformers for near-infinite context}.
\newblock \emph{ArXiv}, abs/2310.01889.

\bibitem[{Mangrulkar et~al.(2022)Mangrulkar, Gugger, Debut, Belkada, Paul, Bossan, and Tietz}]{peft}
Sourab Mangrulkar, Sylvain Gugger, Lysandre Debut, Younes Belkada, Sayak Paul, Benjamin Bossan, and Marian Tietz. 2022.
\newblock {PEFT}: State-of-the-art parameter-efficient fine-tuning methods.
\newblock \url{https://github.com/huggingface/peft}.

\bibitem[{Moritz et~al.(2018)Moritz, Nishihara, Wang, Tumanov, Liaw, Liang, Elibol, Yang, Paul, Jordan, and Stoica}]{moritz2018ray}
Philipp Moritz, Robert Nishihara, Stephanie Wang, Alexey Tumanov, Richard Liaw, Eric Liang, Melih Elibol, Zongheng Yang, William Paul, Michael~I. Jordan, and Ion Stoica. 2018.
\newblock \href {https://www.usenix.org/conference/osdi18/presentation/moritz} {Ray: A distributed framework for emerging {AI} applications}.
\newblock In \emph{13th USENIX Symposium on Operating Systems Design and Implementation (OSDI 18)}, pages 561--577, Carlsbad, CA. USENIX Association.

\bibitem[{Or et~al.(2025)Or, Jain, Vega-Myhre, Cai, Hernandez, Zheng, Guessous, Kuznetsov, Puhrsch, Saroufim, Rao, Tran, and Samardžić}]{torchao}
Andrew Or, Apurva Jain, Daniel Vega-Myhre, Jesse Cai, Charles~David Hernandez, Zhenrui Zheng, Driss Guessous, Vasiliy Kuznetsov, Christian Puhrsch, Mark Saroufim, Supriya Rao, Thien Tran, and Aleksandar Samardžić. 2025.
\newblock \href {https://arxiv.org/abs/2507.16099} {Torchao: Pytorch-native training-to-serving model optimization}.
\newblock \emph{Preprint}, arXiv:2507.16099.

\bibitem[{Ouyang et~al.(2022)Ouyang, Wu, Jiang, Almeida, Wainwright, Mishkin, Zhang, Agarwal, Slama, Ray, Schulman, Hilton, Kelton, Miller, Simens, Askell, Welinder, Christiano, Leike, and Lowe}]{ouyang2022training}
Long Ouyang, Jeffrey Wu, Xu~Jiang, Diogo Almeida, Carroll Wainwright, Pamela Mishkin, Chong Zhang, Sandhini Agarwal, Katarina Slama, Alex Ray, John Schulman, Jacob Hilton, Fraser Kelton, Luke Miller, Maddie Simens, Amanda Askell, Peter Welinder, Paul~F Christiano, Jan Leike, and Ryan Lowe. 2022.
\newblock \href {https://proceedings.neurips.cc/paper_files/paper/2022/file/b1efde53be364a73914f58805a001731-Paper-Conference.pdf} {Training language models to follow instructions with human feedback}.
\newblock In \emph{Advances in Neural Information Processing Systems}, volume~35, pages 27730--27744. Curran Associates, Inc.

\bibitem[{Paszke et~al.(2019)Paszke, Gross, Massa, Lerer, Bradbury, Chanan, Killeen, Lin, Gimelshein, Antiga, Desmaison, Kopf, Yang, DeVito, Raison, Tejani, Chilamkurthy, Steiner, Fang, Bai, and Chintala}]{pytorch}
Adam Paszke, Sam Gross, Francisco Massa, Adam Lerer, James Bradbury, Gregory Chanan, Trevor Killeen, Zeming Lin, Natalia Gimelshein, Luca Antiga, Alban Desmaison, Andreas Kopf, Edward Yang, Zachary DeVito, Martin Raison, Alykhan Tejani, Sasank Chilamkurthy, Benoit Steiner, Lu~Fang, and 2 others. 2019.
\newblock \href {https://proceedings.neurips.cc/paper_files/paper/2019/file/bdbca288fee7f92f2bfa9f7012727740-Paper.pdf} {Pytorch: An imperative style, high-performance deep learning library}.
\newblock In \emph{Advances in Neural Information Processing Systems}, volume~32. Curran Associates, Inc.

\bibitem[{{PyTorch Community}(2023)}]{fsdp-rfc}
{PyTorch Community}. 2023.
\newblock \href {https://github.com/pytorch/pytorch/issues/114299} {{PyTorch FSDP2 RFC}}.
\newblock GitHub Issue.

\bibitem[{{PyTorch Community}(2026)}]{pytorch_contributors_torchdistributedtensor_2026}
{PyTorch Community}. 2026.
\newblock \href {https://docs.pytorch.org/docs/2.12/distributed.tensor.html} {{torch.distributed.tensor --- PyTorch 2.12 Documentation}}.

\bibitem[{{Qwen Team}(2025)}]{yang2025qwen3}
{Qwen Team}. 2025.
\newblock \href {https://arxiv.org/abs/2505.09388} {{Qwen3} technical report}.
\newblock \emph{Preprint}, arXiv:2505.09388.

\bibitem[{Rafailov et~al.(2023)Rafailov, Sharma, Mitchell, Manning, Ermon, and Finn}]{rafailov2023direct}
Rafael Rafailov, Archit Sharma, Eric Mitchell, Christopher~D Manning, Stefano Ermon, and Chelsea Finn. 2023.
\newblock \href {https://proceedings.neurips.cc/paper_files/paper/2023/file/a85b405ed65c6477a4fe8302b5e06ce7-Paper-Conference.pdf} {Direct preference optimization: Your language model is secretly a reward model}.
\newblock In \emph{Advances in Neural Information Processing Systems}, volume~36, pages 53728--53741. Curran Associates, Inc.

\bibitem[{Rajbhandari et~al.(2020)Rajbhandari, Rasley, Ruwase, and He}]{zero9355301}
Samyam Rajbhandari, Jeff Rasley, Olatunji Ruwase, and Yuxiong He. 2020.
\newblock \href {https://doi.org/10.1109/SC41405.2020.00024} {Zero: Memory optimizations toward training trillion parameter models}.
\newblock In \emph{SC20: International Conference for High Performance Computing, Networking, Storage and Analysis}, pages 1--16.

\bibitem[{Rasley et~al.(2020)Rasley, Rajbhandari, Ruwase, and He}]{deepspeed}
Jeff Rasley, Samyam Rajbhandari, Olatunji Ruwase, and Yuxiong He. 2020.
\newblock \href {https://doi.org/10.1145/3394486.3406703} {Deepspeed: System optimizations enable training deep learning models with over 100 billion parameters}.
\newblock In \emph{Proceedings of the 26th ACM SIGKDD International Conference on Knowledge Discovery \& Data Mining}, KDD '20, page 3505–3506, New York, NY, USA. Association for Computing Machinery.

\bibitem[{Schulman et~al.(2017)Schulman, Wolski, Dhariwal, Radford, and Klimov}]{schulman2017proximal}
John Schulman, Filip Wolski, Prafulla Dhariwal, Alec Radford, and Oleg Klimov. 2017.
\newblock \href {https://arxiv.org/abs/1707.06347} {Proximal policy optimization algorithms}.
\newblock \emph{Preprint}, arXiv:1707.06347.

\bibitem[{Shao et~al.(2024)Shao, Wang, Zhu, Xu, Song, Bi, Zhang, Zhang, Li, Wu, and Guo}]{shao2024deepseekmath}
Zhihong Shao, Peiyi Wang, Qihao Zhu, Runxin Xu, Junxiao Song, Xiao Bi, Haowei Zhang, Mingchuan Zhang, Y.~K. Li, Y.~Wu, and Daya Guo. 2024.
\newblock \href {https://arxiv.org/abs/2402.03300} {Deepseekmath: Pushing the limits of mathematical reasoning in open language models}.
\newblock \emph{Preprint}, arXiv:2402.03300.

\bibitem[{Tillet et~al.(2019)Tillet, Kung, and Cox}]{triton}
Philippe Tillet, H.~T. Kung, and David Cox. 2019.
\newblock \href {https://doi.org/10.1145/3315508.3329973} {Triton: an intermediate language and compiler for tiled neural network computations}.
\newblock In \emph{Proceedings of the 3rd ACM SIGPLAN International Workshop on Machine Learning and Programming Languages}, MAPL 2019, page 10–19, New York, NY, USA. Association for Computing Machinery.

\bibitem[{von Werra et~al.(2020)von Werra, Belkada, Tunstall, Beeching, Thrush, Lambert, Huang, Rasul, and Gallouédec}]{vonwerra2022trl}
Leandro von Werra, Younes Belkada, Lewis Tunstall, Edward Beeching, Tristan Thrush, Nathan Lambert, Shengyi Huang, Kashif Rasul, and Quentin Gallouédec. 2020.
\newblock Trl: Transformers reinforcement learning.
\newblock \url{https://github.com/huggingface/trl}.

\bibitem[{Wijmans et~al.(2025)Wijmans, Huval, Hertzberg, Koltun, and Kraehenbuehl}]{wijmans2025cut}
Erik Wijmans, Brody Huval, Alexander Hertzberg, Vladlen Koltun, and Philipp Kraehenbuehl. 2025.
\newblock \href {https://openreview.net/forum?id=E4Fk3YuG56} {Cut your losses in large-vocabulary language models}.
\newblock In \emph{The Thirteenth International Conference on Learning Representations}.

\bibitem[{Wolf et~al.(2020)Wolf, Debut, Sanh, Chaumond, Delangue, Moi, Cistac, Rault, Louf, Funtowicz, Davison, Shleifer, von Platen, Ma, Jernite, Plu, Xu, Le~Scao, Gugger, Drame, Lhoest, and Rush}]{wolf-etal-2020-transformers}
Thomas Wolf, Lysandre Debut, Victor Sanh, Julien Chaumond, Clement Delangue, Anthony Moi, Pierric Cistac, Tim Rault, Remi Louf, Morgan Funtowicz, Joe Davison, Sam Shleifer, Patrick von Platen, Clara Ma, Yacine Jernite, Julien Plu, Canwen Xu, Teven Le~Scao, Sylvain Gugger, and 3 others. 2020.
\newblock \href {https://doi.org/10.18653/v1/2020.emnlp-demos.6} {Transformers: State-of-the-art natural language processing}.
\newblock In \emph{Proceedings of the 2020 Conference on Empirical Methods in Natural Language Processing: System Demonstrations}, pages 38--45, Online. Association for Computational Linguistics.

\bibitem[{Yadan(2019)}]{Yadan2019Hydra}
Omry Yadan. 2019.
\newblock \href {https://github.com/facebookresearch/hydra} {Hydra - a framework for elegantly configuring complex applications}.
\newblock Github.

\bibitem[{Zhao et~al.(2023)Zhao, Gu, Varma, Luo, Huang, Xu, Wright, Shojanazeri, Ott, Shleifer, Desmaison, Balioglu, Damania, Nguyen, Chauhan, Hao, Mathews, and Li}]{zhao2023pytorchfsdpexperiencesscaling}
Yanli Zhao, Andrew Gu, Rohan Varma, Liang Luo, Chien-Chin Huang, Min Xu, Less Wright, Hamid Shojanazeri, Myle Ott, Sam Shleifer, Alban Desmaison, Can Balioglu, Pritam Damania, Bernard Nguyen, Geeta Chauhan, Yuchen Hao, Ajit Mathews, and Shen Li. 2023.
\newblock \href {https://arxiv.org/abs/2304.11277} {Pytorch fsdp: Experiences on scaling fully sharded data parallel}.
\newblock \emph{Preprint}, arXiv:2304.11277.

\end{thebibliography}

\appendix
\onecolumn

\section{Context parallel}

\texttt{torchtune}'s context-parallel mechanism makes it possible to run post-training on datasets that require context windows of more than 1 million tokens. To evaluate our context-parallel integration, we created a synthetic dataset by concatenating Alpaca samples, with most examples reaching approximately 1M tokens. The dataset-generation code is provided in the supplementary material. In this setup, the \textit{Llama~3.2} model was post-trained on the synthetic dataset with a batch size of 1 on an 8\,$\times$\,H100 cluster. Peak memory allocation occurred during loss computation, reaching 79.2\,GB of VRAM per GPU. The average throughput was 6,720 tokens/s. During the experiment, the following memory trace was logged, showing that the peak memory is represented by the largest of these communication-phase allocations, which remains stable across ring iterations and does not grow with sequence length beyond the local chunk size.

\begin{figure}[h]
    \centering
    \includegraphics[width=1.0\linewidth]{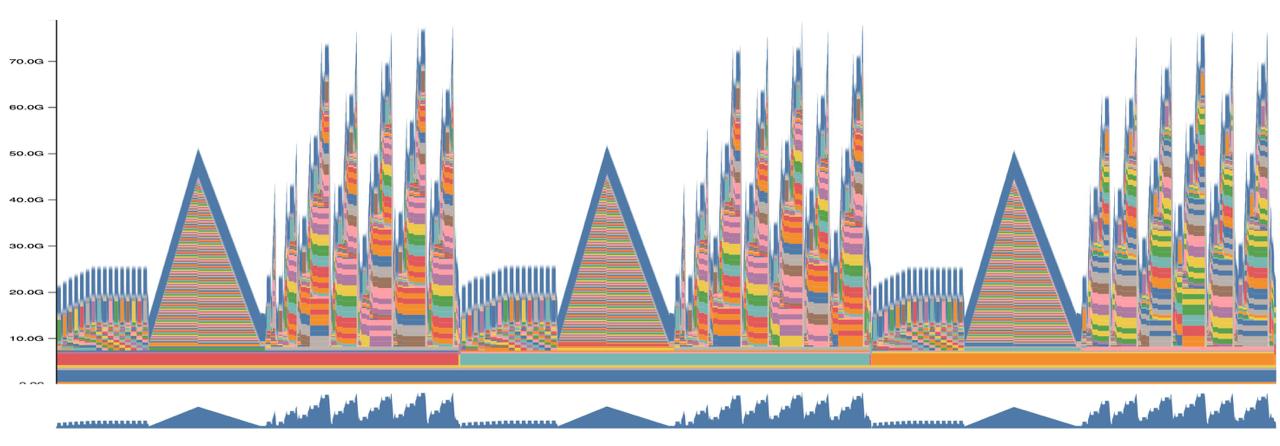}
    \caption{Context Parallel memory trace}
    \label{fig:placeholder}
\end{figure}

\section{Supported Models \& Datasets}
\begin{table}[!htbp]
\centering
\small
\caption{Supported foundation model families and available parameter scales (B = billions of parameters; E = experts for MoE models).}
\begin{tabular}{@{}ll@{}}
\toprule
\textbf{Model} & \textbf{Sizes} \\
\midrule
Llama~4          & Scout (17B$\times$16E) \\
Llama~3.3        & 70B \\
Llama~3.2-Vision & 11B, 90B \\
Llama~3.2        & 1B, 3B \\
Llama~3.1        & 8B, 70B, 405B \\
Mistral          & 7B \\
Gemma~2          & 2B, 9B, 27B \\
Microsoft Phi-4  & 14B \\
Microsoft Phi-3  & Mini \\
Qwen~3           & 0.6B, 1.7B, 4B, 8B, 14B, 30B-A3B, 32B, 235B-A22B \\
Qwen~2.5         & 0.5B, 1.5B, 3B, 7B, 14B, 32B, 72B \\
Qwen~2           & 0.5B, 1.5B, 7B \\
\bottomrule
\end{tabular}
\label{tab:models_support}
\end{table}

\begin{table*}[h!]
\centering
\footnotesize
\setlength{\tabcolsep}{6pt}
\renewcommand{\arraystretch}{1.15}
\caption{Supported dataset types and built-in preset dataset builders in torchtune.}
\begin{tabularx}{\textwidth}{@{}l l >{\raggedright\arraybackslash}X@{}}
\toprule
\textbf{Type} & \textbf{Entry point} & \textbf{Schema + built-in presets} \\
\midrule
Chat & \texttt{chat\_dataset} &
Conversation (\texttt{conversations/messages}): \texttt{slimorca\_dataset}. \\

Instruct & \texttt{instruct\_dataset} &
Input/output pairs: \texttt{alpaca\_dataset}, \texttt{grammar\_dataset}, \texttt{samsum\_dataset}. \\

Multimodal & \texttt{multimodal\_chat\_dataset} &
Chat + image-path: \texttt{the\_cauldron\_dataset}, \texttt{llava\_instruct\_dataset}. \\

Preference (DPO) & \texttt{preference\_dataset} &
Chosen/rejected: \texttt{hh\_rlhf\_helpful\_dataset}, \texttt{stack\_exchange\_paired\_dataset}. \\

Text completion & \texttt{text\_completion\_dataset} &
Single text column: \texttt{cnn\_dailymail\_articles\_dataset}. \\
\bottomrule
\end{tabularx}
\label{tab:datasets_support}
\end{table*}

\newpage

\section{Component and builder examples}
\label{app:component_examples}

The excerpts below illustrate the component and builder pattern discussed in \cref{sec:components_builders}. The decoder receives already constructed modules, the attention block receives externally supplied projections, and builders assemble the concrete module tree for a model family.

\begin{minted}[
    breaklines,
    breakanywhere,
    fontsize=\footnotesize
]{python}
class TransformerDecoder(nn.Module):
    def __init__(
        self,
        *,
        tok_embeddings,
        layers,
        max_seq_len,
        ...
    ) -> None:
\end{minted}

\begin{minted}[
    breaklines,
    breakanywhere,
    fontsize=\footnotesize
]{python}
class MultiHeadAttention(nn.Module):
    def __init__(
        self,
        *,
        embed_dim: int,
        num_heads: int,
        num_kv_heads: int,
        head_dim: int,
        q_proj: nn.Module,
        k_proj: nn.Module,
        v_proj: nn.Module,
        output_proj: nn.Module,
        ...
    ) -> None:
\end{minted}

A builder exposes the architectural hyperparameters and assembles the layer graph explicitly.

\begin{minted}[
    breaklines,
    breakanywhere,
    fontsize=\footnotesize
]{python}
def llama3(
    vocab_size: int,
    num_layers: int,
    num_heads: int,
    num_kv_heads: int,
    embed_dim: int,
    max_seq_len: int,
    attn_dropout: float = 0.0,
    rope_base: int = 500_000,
    intermediate_dim: Optional[int] = None,
    norm_eps: float = 1e-5,
) -> TransformerDecoder:
\end{minted}

\begin{minted}[
    breaklines,
    breakanywhere,
    fontsize=\footnotesize
]{python}
layers = nn.ModuleList()
for _ in range(num_layers):
    self_attn = MultiHeadAttention(
        embed_dim=embed_dim,
        num_heads=num_heads,
        num_kv_heads=num_kv_heads,
        head_dim=head_dim,
        q_proj=nn.Linear(...),
        k_proj=nn.Linear(...),
        v_proj=nn.Linear(...),
        output_proj=nn.Linear(...),
        pos_embeddings=rope,
        max_seq_len=max_seq_len,
        attn_dropout=attn_dropout,
    )
    mlp = llama3_mlp(dim=embed_dim, hidden_dim=hidden_dim)
    layer = TransformerSelfAttentionLayer(
        attn=self_attn,
        mlp=mlp,
        sa_norm=RMSNorm(...),
        mlp_norm=RMSNorm(...),
    )
    layers.append(layer)
\end{minted}

\begin{minted}[
    breaklines,
    breakanywhere,
    fontsize=\footnotesize
]{python}
return TransformerDecoder(
    tok_embeddings=tok_embeddings,
    layers=layers,
    max_seq_len=max_seq_len,
    num_heads=num_heads,
    head_dim=head_dim,
    norm=RMSNorm(embed_dim, eps=norm_eps),
    output=output_proj,
)
\end{minted}

The same pattern supports parameter-efficient fine-tuning. For LoRA~\citep{hu_lora_2022}, the builder substitutes low-rank projection modules while leaving the shared attention implementation unchanged.

\begin{minted}[
    breaklines,
    breakanywhere,
    fontsize=\footnotesize
]{python}
self_attn = MultiHeadAttention(
    embed_dim=embed_dim,
    num_heads=num_heads,
    num_kv_heads=num_kv_heads,
    head_dim=head_dim,
    q_proj=LoRALinear(embed_dim, ...),
    k_proj=LoRALinear(embed_dim, ...),
    v_proj=LoRALinear(embed_dim, ...),
    output_proj=nn.Linear(embed_dim, ...),
    ...
)
\end{minted}

\clearpage

\section{Example config}
\label{app:config}

The excerpt below shows the structure of a typical \texttt{torchtune} recipe configuration. Each top-level block names either a component to instantiate or a runtime policy to apply, and the same fields can be replaced in YAML or overridden from the command line for ablations and sweeps.

\begin{minted}[
    breaklines,
    breakanywhere,
    fontsize=\footnotesize
]{yaml}

model:
  _component_: ...

checkpointer:
  ...
  
resume_from_checkpoint: False

tokenizer:
  _component_: ...
  path: ...
  max_seq_len: null

dataset:
  _component_: ...
  packed: False  
  split: train[:95%]
seed: null
shuffle: True

run_val_every_n_steps: null  
dataset_val:
  _component_: ...
  split: train[95%:]
batch_size_val: ${batch_size}

batch_size: 2
epochs: 1
optimizer:
  _component_: torch.optim.AdamW
  fused: True
  lr: 2e-5
loss:
  _component_: ...
max_steps_per_epoch: null
gradient_accumulation_steps: 1  
clip_grad_norm: null
compile: False 
optimizer_in_bwd: False 

device: cuda

enable_activation_checkpointing: True 
enable_activation_offloading: False  

dtype: bf16

metric_logger:
  _component_: ...
  log_dir: ${output_dir}/logs
log_every_n_steps: 1
log_peak_memory_stats: True
log_level: INFO 
profiler:
  _component_: ...
  enabled: False

  ...
\end{minted}

\FloatBarrier

\end{document}